\documentclass[conference]{IEEEtran}
\IEEEoverridecommandlockouts
\usepackage{cite}
\usepackage{amsmath,amssymb,amsfonts}
\usepackage{algorithmic}
\usepackage{graphicx}
\usepackage{textcomp}
\usepackage{xcolor}
\usepackage{listings}
\usepackage{float}
\def\BibTeX{{\rm B\kern-.05em{\sc i\kern-.025em b}\kern-.08em
    T\kern-.1667em\lower.7ex\hbox{E}\kern-.125emX}}
\begin{document}

\title{Real-time Object Detection: \\ YOLOv1 Re-Implementation in PyTorch}

\author{
\IEEEauthorblockN{Michael Shenoda}
\IEEEauthorblockA{
\textit{College of Computing \& Informatics} \\
\textit{Drexel University}\\
Philadelphia, PA, United States \\
michael.shenoda@drexel.edu \\
michael@mshenoda.com
}}

\maketitle

\begin{abstract}
Real-time object detection is a crucial problem to solve when in comes to computer vision systems that needs to make appropriate decision based on detection in a timely manner. 
I have chosen the YOLO v1 architecture to implement it using PyTorch framework, with goal to familiarize myself with entire object detection pipeline 
I attempted different techniques to modify the original architecture to improve the results. 
Finally, I compare the metrics of my implementation to the original.
\end{abstract}

\section{Introduction}
Surprisingly humans are really good at understanding visual content of an image and instantly provide information about the objects within.
Since 2012 Convolutional Neural Network (CNN) has became so popular for object detection and classification tasks, but yet the challenges to create reliable models can get overwhelming.
Looking at the YOLO architecture, it has taken a leap forward as being the architecture of choice for real-time detectors. 
I have chosen the YOLO v1 architecture to implement with primary goal to learn the fundamentals. 
Note that the original YOLO is implemented in Darknet c framework. So, the goal is to familiarize myself with object detection training and inference pipeline and understand the proper way to implement that using PyTorch framework. 
I attempted different techniques to modify the original architecture such as changes to kernel sizes as well as network depth and changing activation layers to improve the results. 
I have created a baseline using YOLO tiny and proposed modified version of the model to attempt to improve the results overall.
Finally, I show the metrics comparing it to baseline and to published YOLO model metrics. 
I also provide a visual validaiton for the detection to visualize how the model is behaviing.
\section{Fundamentals of YOLO}
\subsection{Dataset}
Starting off with the dataset, YOLO used the PASCAL Visual Object Classes dataset, the 2007 and 2012 combined.
It’s important to note that a pre-processing step is required to convert the labels from VOC format to YOLO format.
The author of YOLO has provided a conversion python script, which is available on GitHub \cite{b2}

\begin{figure}[H]
  \centering
  \includegraphics[width=\linewidth]{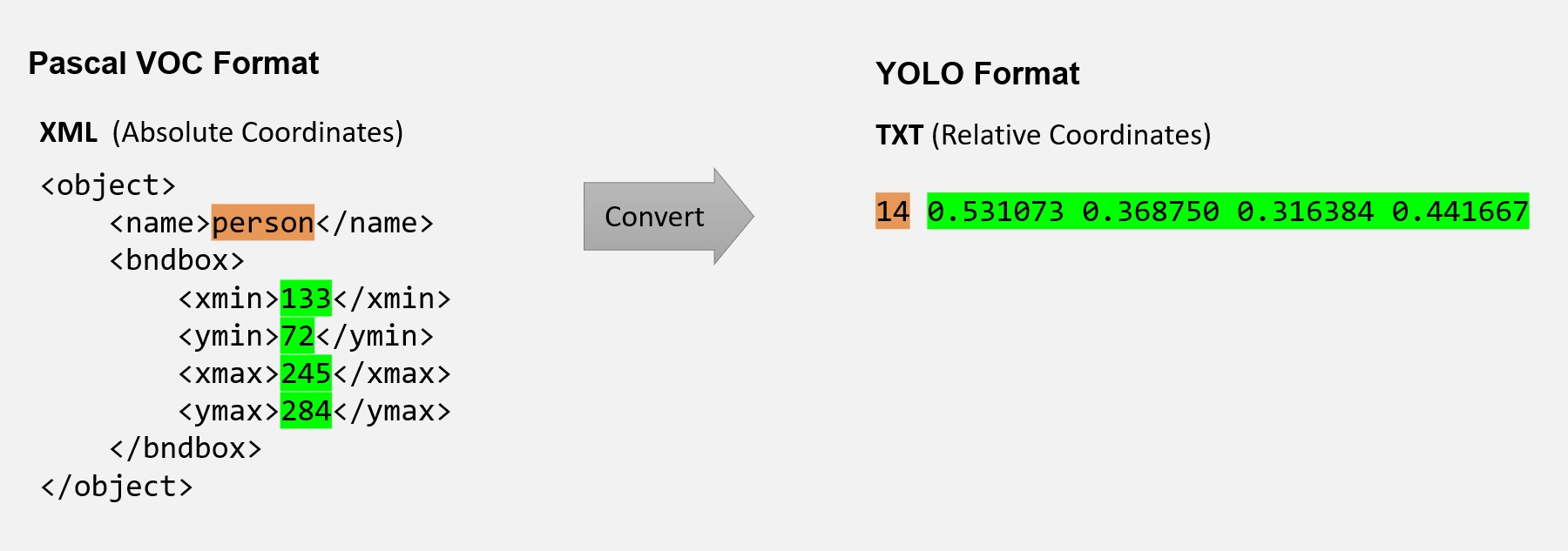}
  \caption{Conversion from VOC to YOLO}
\end{figure}

\subsection{Bounding Box Labels}
The YOLO bounding box label starts off by a numerical class id, followed by normalized box coordinates between 0 to 1
It’s important to note that the x, y coordinate here is the bounding box center.
They have chosen that to allow the bounding box to be scalable regardless of the image size. 
\begin{figure}[H]
  \centering
  \includegraphics[width=\linewidth]{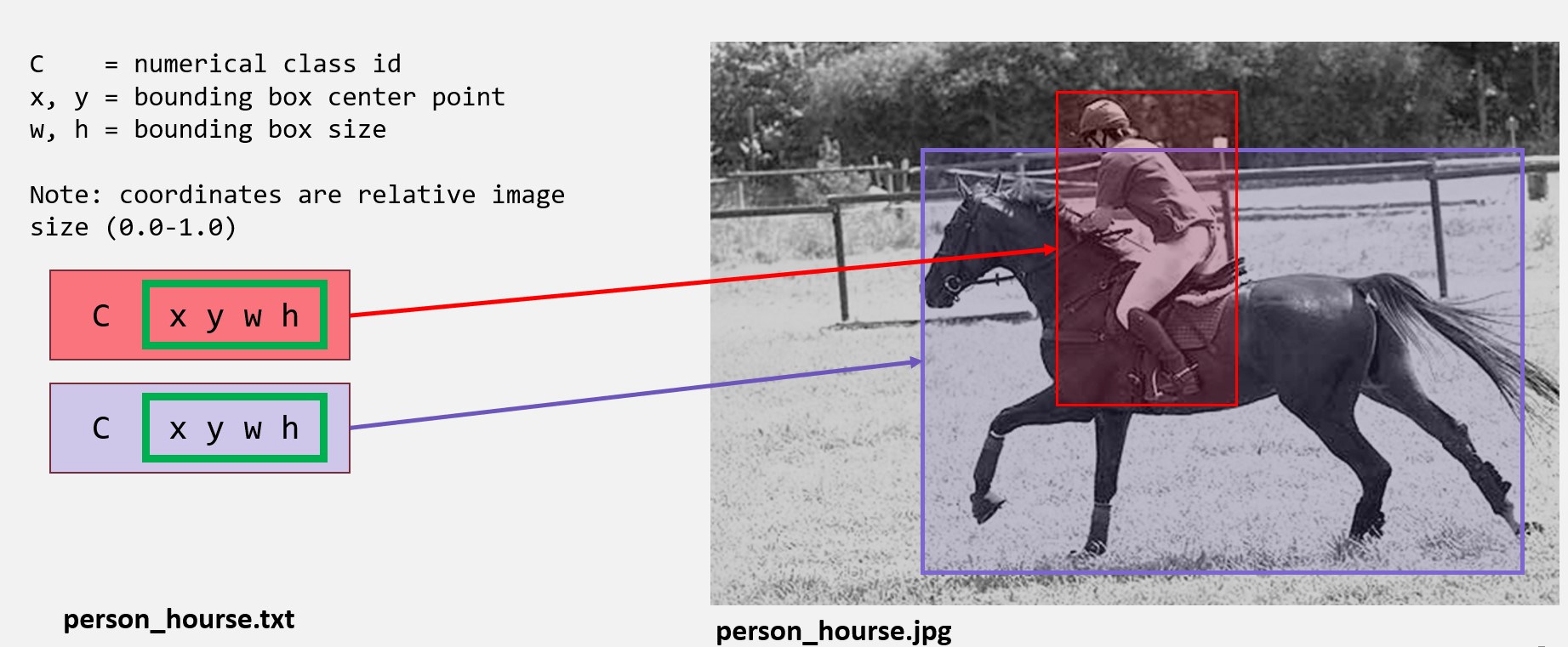}
  \caption{Bounding Box Labels Per Image}
\end{figure}
\subsection{Tensor Structure}
Before diving into the model architecture, lets look at the tensor structure.
Looking at the tensor from a 2D the surface, it’s a grid divided by the chosen grid size of 7x7
If we look at the tensor depth, each cell contains two bounding boxes followed by an object pretense flag, then followed by class probabilities. 
The bounding boxes here are relative to the cell, instead of the whole image. So, an encoding step is required to encode the raw label into a tensor before training.
Also, a decoding step is required while doing inference, to decode the tensor back to a box coordinate relative to the image and get the class information of the object.
\begin{figure}[H]
  \centering
  \includegraphics[width=\linewidth]{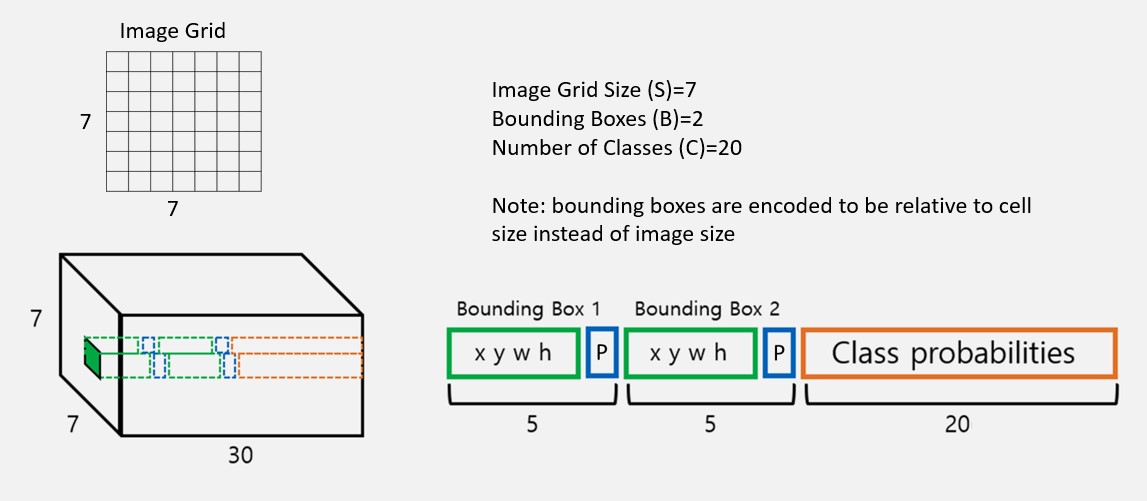}
  \caption{YOLO Tensor Structure}
\end{figure}

\subsection{Full Architecture}
Yolo has 24 convolutional layers, alternating 1 × 1 convolutional layers reduce the features space from preceding layers, then followed by 2 fully connected layers to produce the final tensor as shown previously.
\begin{figure}[H]
  \centering
  \includegraphics[width=\linewidth]{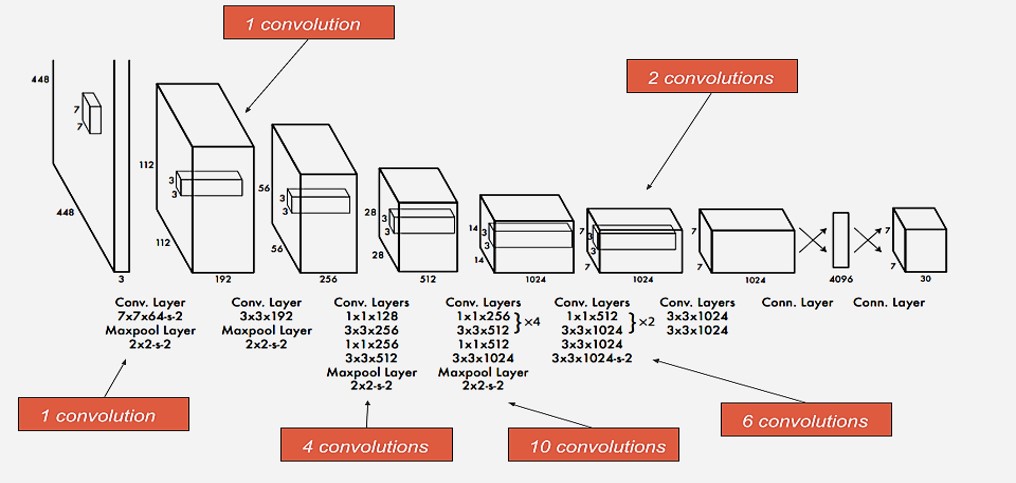}
  \caption{YOLO Full Model Architecture}
\end{figure}
\subsection{Tiny Baseline Architecture}
For my baseline, I chosen the tiny architecture of YOLO to implement, due to time and resources.
The tiny architecture composed of 9 convolutional layers followed by 2 fully connected to produce the final tensor. 
Note here that I reduced the fully connected size from 4096 to 2048 to reduce memory footprint. 
\begin{figure}[H]
  \centering
  \includegraphics[width=\linewidth]{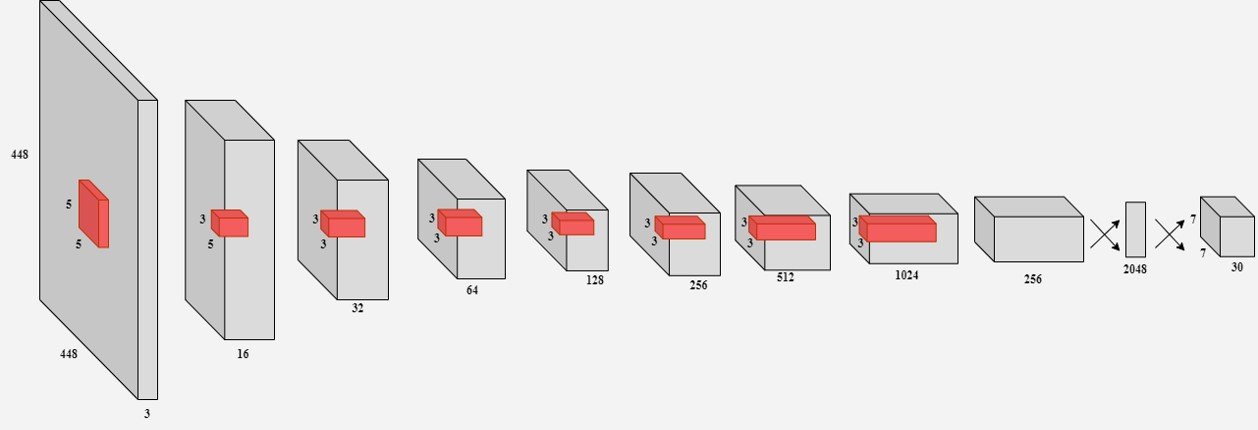}
  \caption{YOLO Tiny Model Baseline Architecture}
\end{figure}
\subsection{MS Architecture}
My modified YOLO model which consist of 6 convolutional layers with some modification to kernel size, activation, 
and adding adaptive average pooling \cite{b5} before fully connected layer. I also reduced the fully connected size from original 4096 down to 1920.
In the Convolutional Layers I have made the following changes:\\
\begin{enumerate}
  \item Used 5x5 kernel size across the board
  \item Replaced first 3 LeakyReLU with ReLU
  \item Replaced 4th \& 5th LeakyReLU with SiLU \cite{b6}
  \item Then added AdaptiveAvgPool2d before Fully Connected
\end{enumerate}

Regarding the FC Layers, I have made the following changes:
\begin{enumerate}
  \item Replaced LeakyReLU with SiLU activation
  \item Changed Dropout of 0.25 after first FC
\end{enumerate}

\begin{figure}[H]
  \centering
  \includegraphics[width=\linewidth]{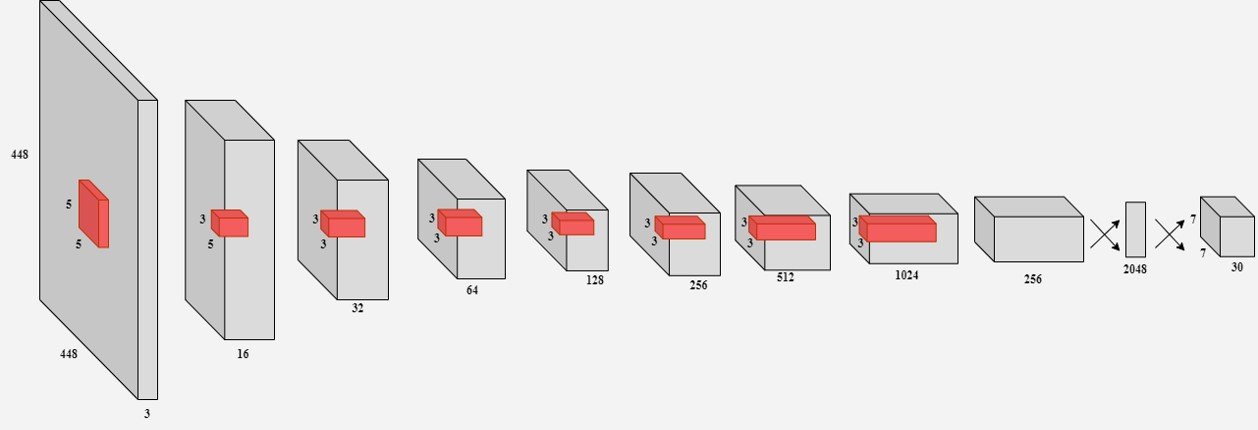}
  \caption{YOLO MS Model Architecture}
\end{figure}
\subsection{Training}
The training of original yolo has been done with 135 epochs with convolutional layers pre-trained on ImageNet at half resolution then doubled up for detection. They mention in the paper that the pertaining took them a week to complete, which I don’t have the time nor resources to do so! Instead, I train everything from scratch with 200 epochs. The baseline used a fixed learning schedule. In my modified yolo, I did two different experiment one with two step fixed learning rate, and the other I used OneCycle learning rate with cosine annealing. 
Everything was trained with same optimizer with momentum and decay. 
\begin{figure}[H]
  \centering
  \includegraphics[width=\linewidth]{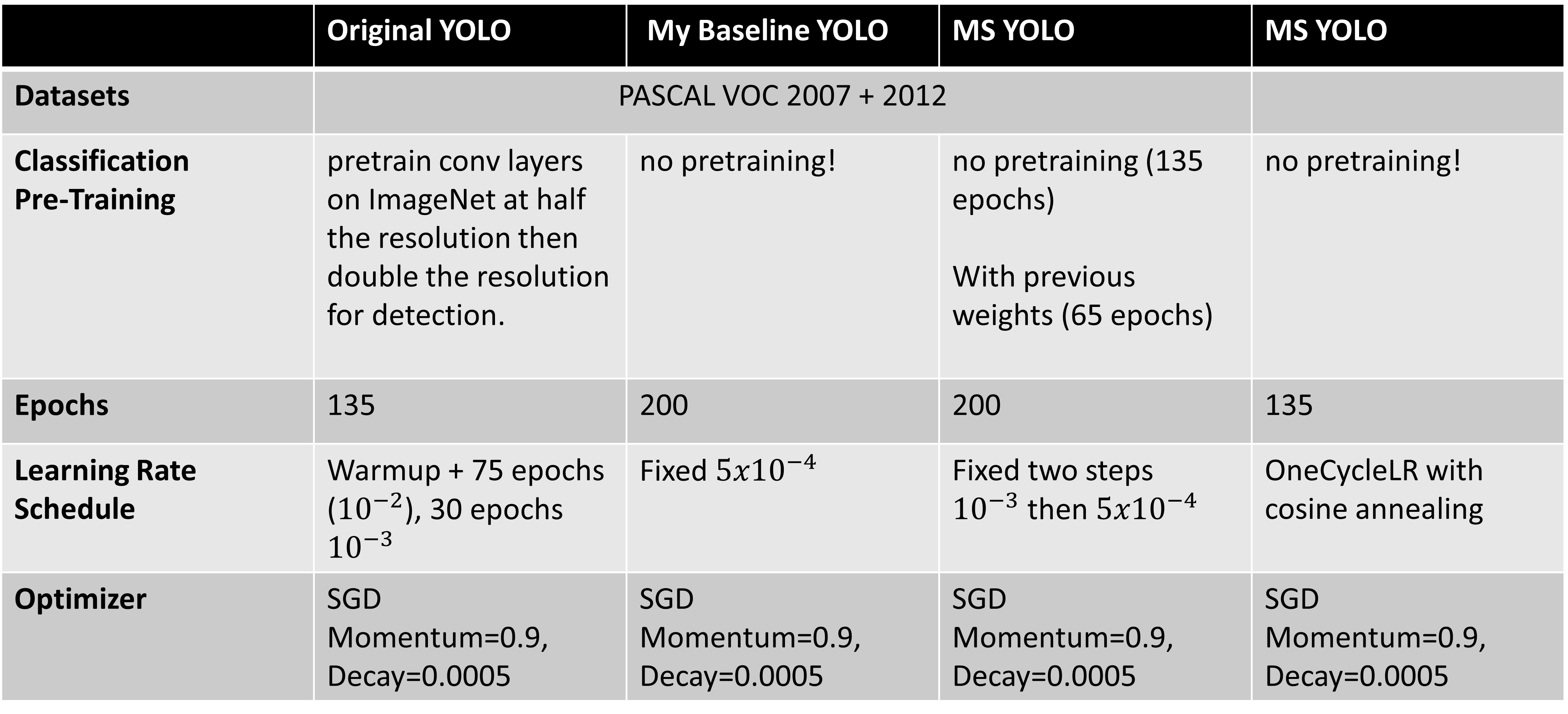}
  \caption{Training Paramters Table}
\end{figure}

\subsection{YOLO Loss}
Now, let’s look at the YOLO loss.
As we can see in the foolwing equation, it's not an off the shelf loss function. 
At first, it may look complicated, but once breaking it into parts it starts to make more sense.
\begin{figure}[H]
  \centering
  \includegraphics[width=\linewidth]{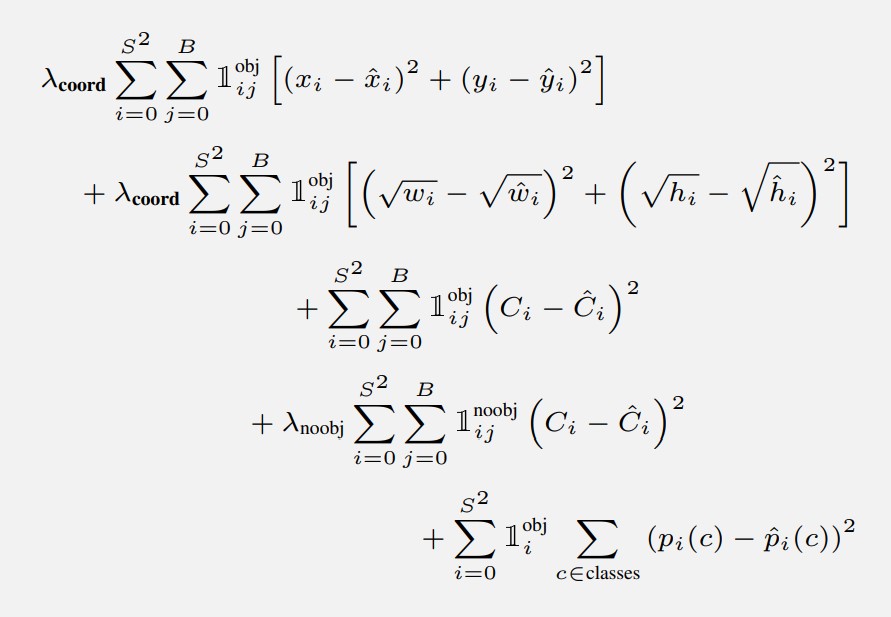}
  \caption{YOLO Loss Function}
\end{figure}

Looking back to the tensor structure mentioned earlier, we will see that it maps nicely with the loss function.
\begin{figure}[H]
  \centering
  \includegraphics[width=\linewidth]{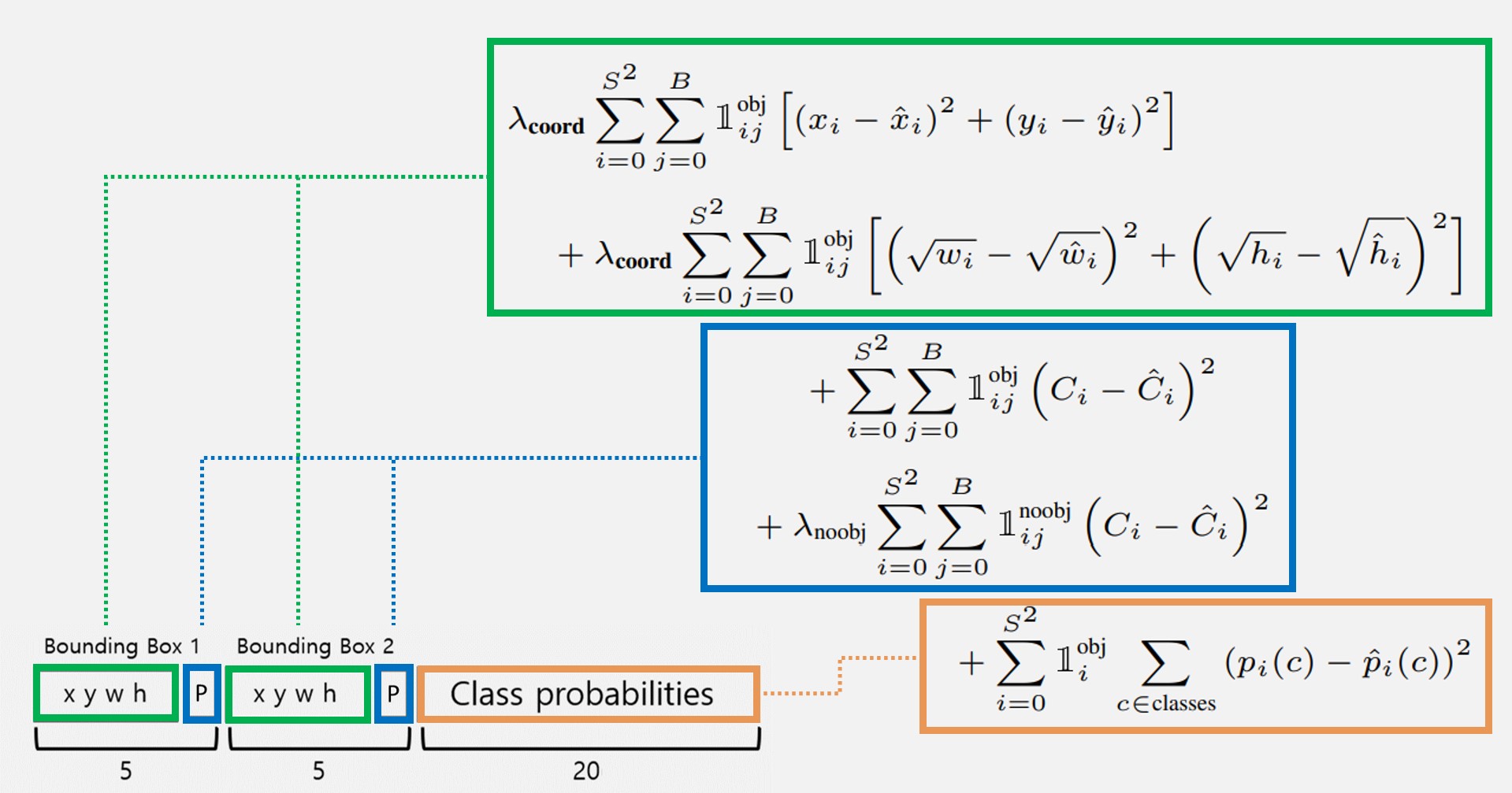}
  \caption{YOLO Loss Function with Visual Marking}
\end{figure}

First part of the equation, highlighted in green, is the coordinate loss which is responsible for the bounding boxes
Second part of the equation, highlighted in blue, is the object presence loss which is responsible for whether the cell contains an object or not! 
Third and last part of the equation, is the object classification loss which is responsible for the class probabilities
Now we can see clearly how the loss is structured and mapped to the tensor structure.
\subsection{YOLO Augmentation}
The original yolo uses color jitter which randomly adjusts brightness, contrast, saturation, and hue. 
It also does a random up scaling of the image up to 20\%.

\section{PyTorch Implementation}
\subsection{Custom Augmentation}
In my implementation, I used color jitter similar to the original, in addition to random blur, random grayscale, random horizontal flip, random vertical flip, random rotation jitter.

\noindent
$ColorJitter(0.2, 0.5, 0.7, 0.07),\\
RandomBlur([3,3], sigma=[0.1, 2], p=0.1),\\
RandomGrayscale(p=0.1),\\
RandomHorizontalFlip(0.5),\\
RandomVerticalFlip(0.05),\\
RandomRotationJitter(p=0.5)\\
$

It’s important to note here that the \textit{RandomBlur, RandomHorizontalFlip, RandomVerticalFlip, and RandomRotationJitter} are custom transform modules that I implemented myself.
Specifically for the last three transforms, they change the location of the pixels, so you cannot just use PyTorch’s off the shelf transforms without taking care of transforming the bounding boxes as well! Yeah! nothing is free when it comes to object detection!

\subsection{Modules Overview}
Below is an overview of the python implementation scripts and configuration.

\begin{figure}[H]
  \centering
  \includegraphics[width=\linewidth]{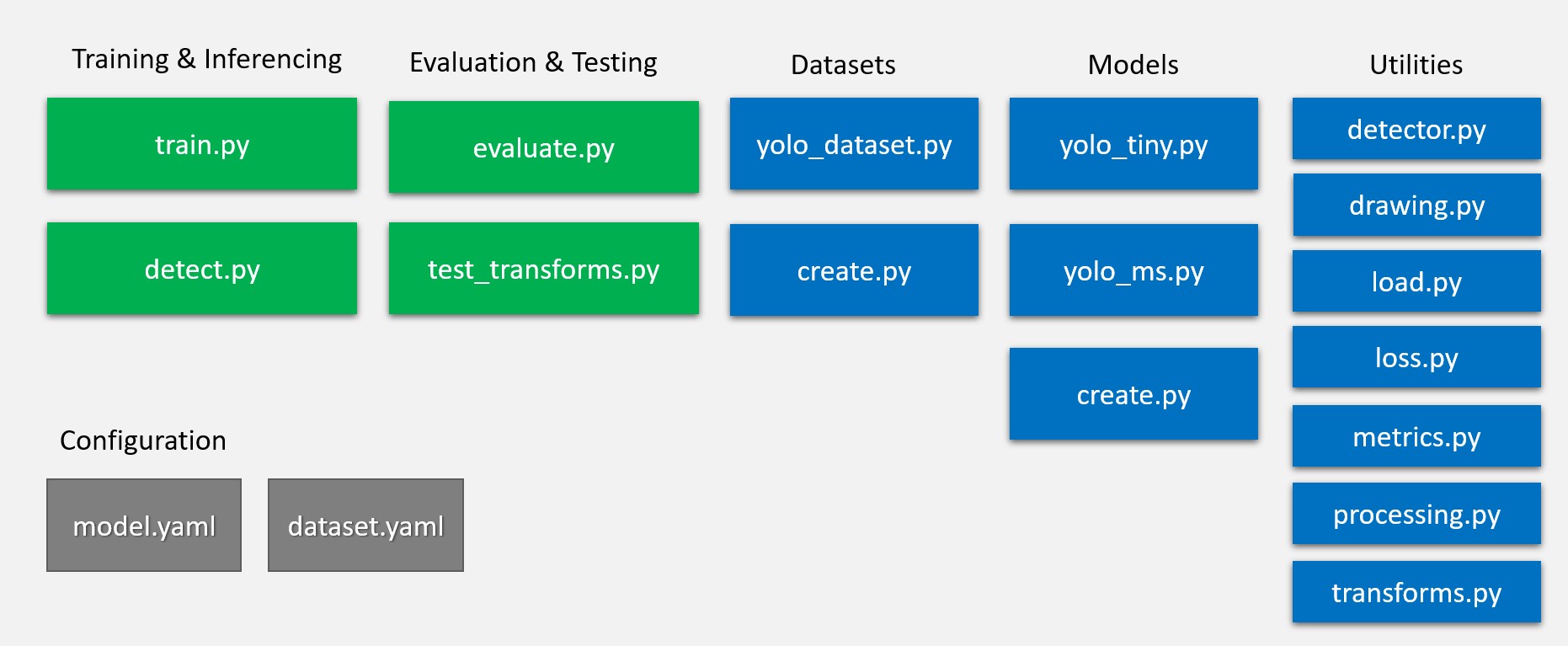}
  \caption{Implementation Modules}
\end{figure}

The ones highlighted in green are the user facing scripts, where you can use to training/evaluation and detection.
I also provide test\_transforms script to easily visualize the effect of image augmentation transforms to help choose the appropriate augmentation for your own dataset. 
The modules highlighted in blue are internal and not intended to be called directly.
\section{Results}
\subsection{Training \& Validation Loss}
Showing below the training and validation loss for the experiments for baseline model and my modified model. 
\begin{figure}[H]
  \centering
  \includegraphics[width=\linewidth]{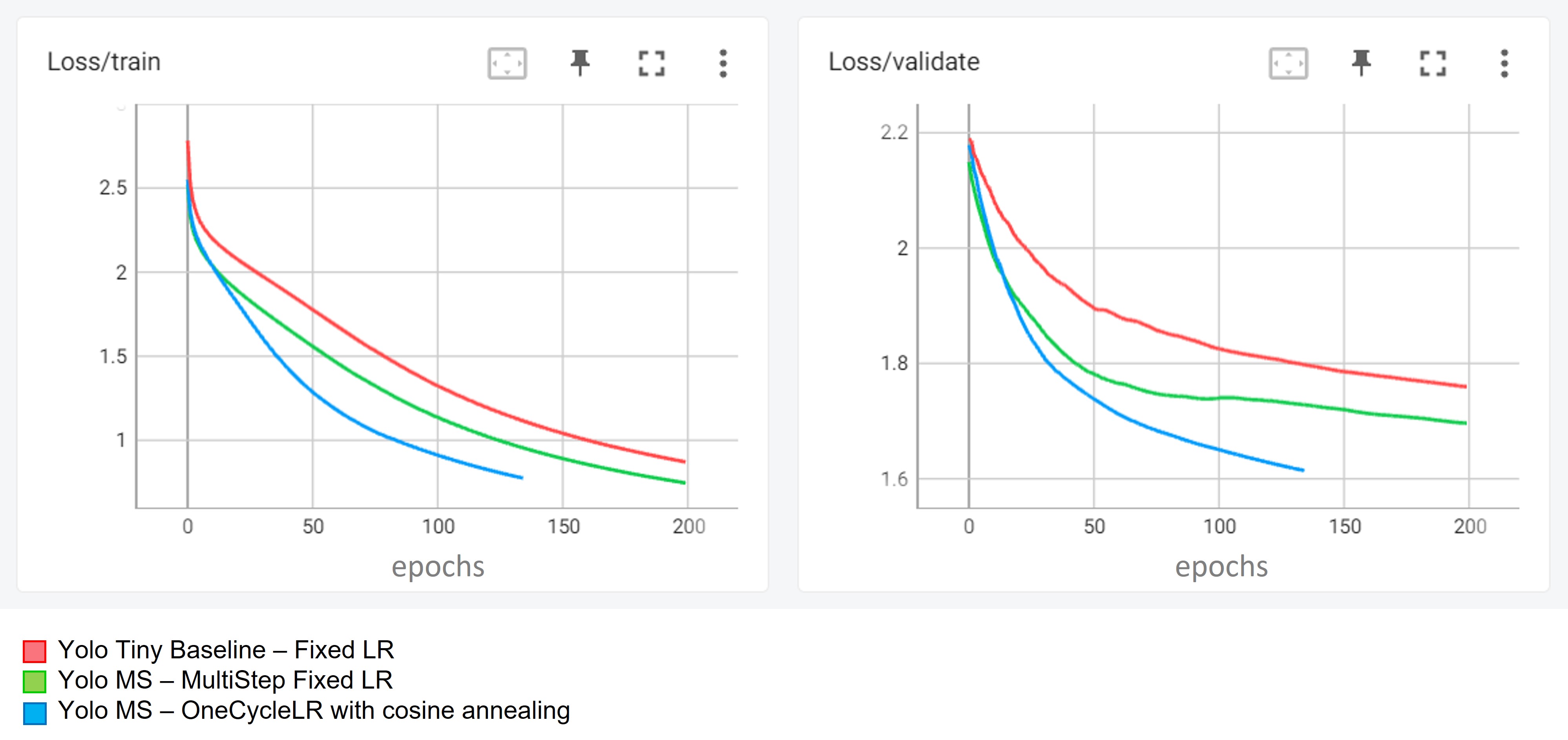}
  \caption{Training and Validation Loss}
\end{figure}
It’s important to note that using OnceCycle \cite{b4} learning schedule with cosine annealing has helped to train and converge faster compared to fixed and multi step fixed learning rate.

\subsection*{Mean Average Precision}
Here is the comparison of the mean average precision for all the models.
The two on the top are my own and the two on the bottom are the published ones. 
\begin{figure}[H]
  \centering
  \includegraphics[width=\linewidth]{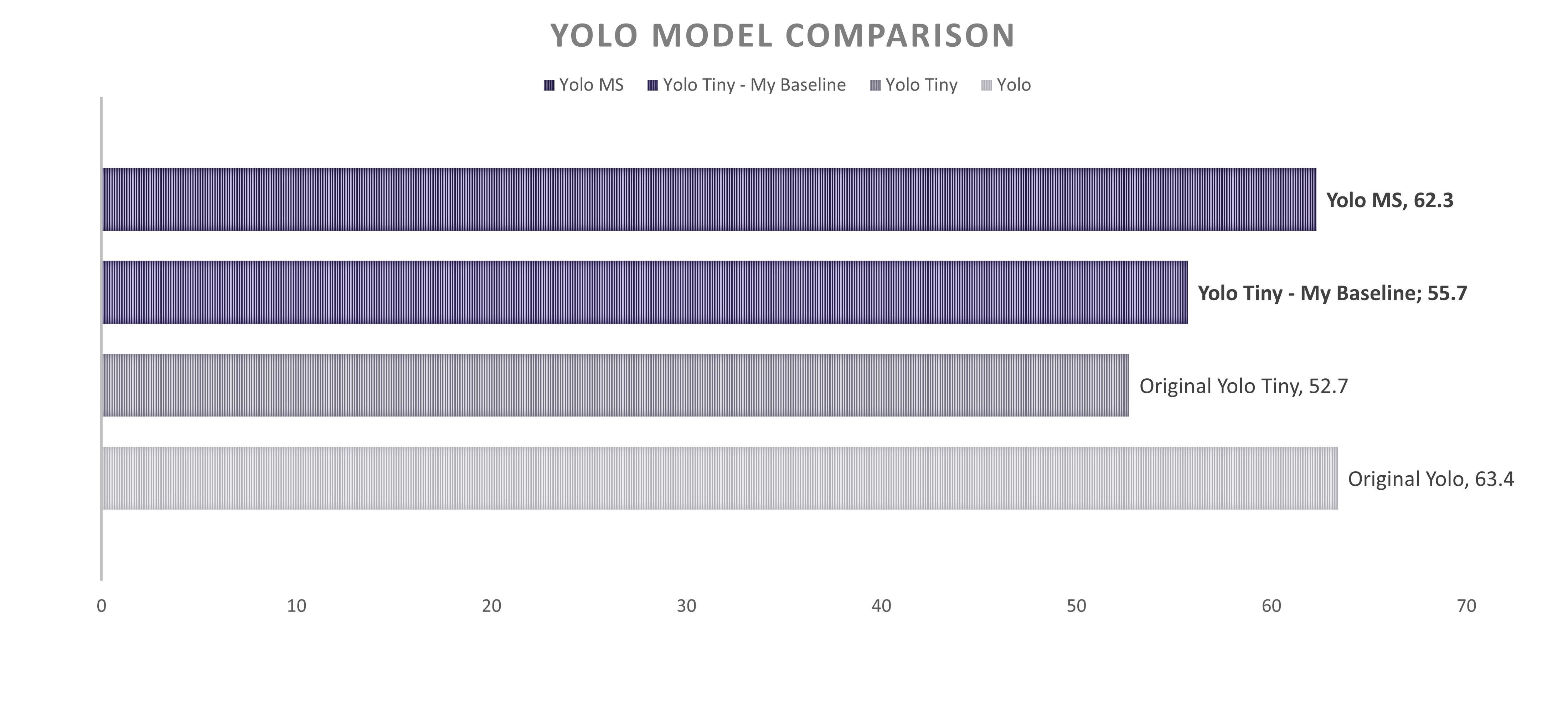}
  \caption{Mean Average Precision Comparison}
\end{figure}
Clearly my latest model at 62.3 mAP is better than Yolo Tiny baseline model. 
It’s getting closer to the result of the full YOLO model but it's super tiny!
An extensive testing would still need to be done to fully confirm the performance. 

\section*{Visual Inspection}
By visually inspecting the detections, we can see that it provides good localization and handles multi object scenarios shown below:
\begin{figure}[H]
  \centering
  \includegraphics[width=\linewidth]{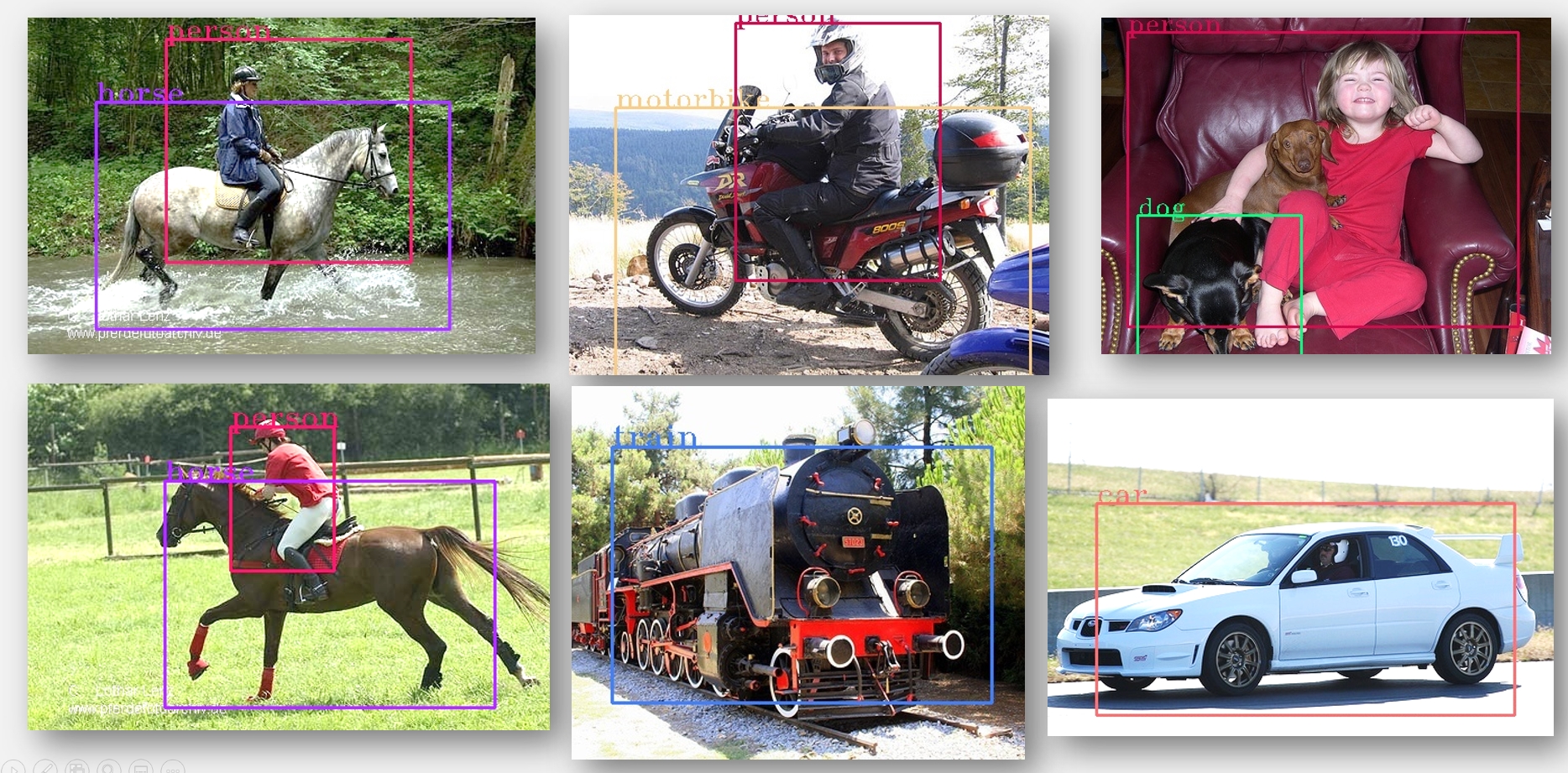}
  \caption{Visual Inspection Part 1}
\end{figure}
\begin{figure}[H]
  \centering
  \includegraphics[width=\linewidth]{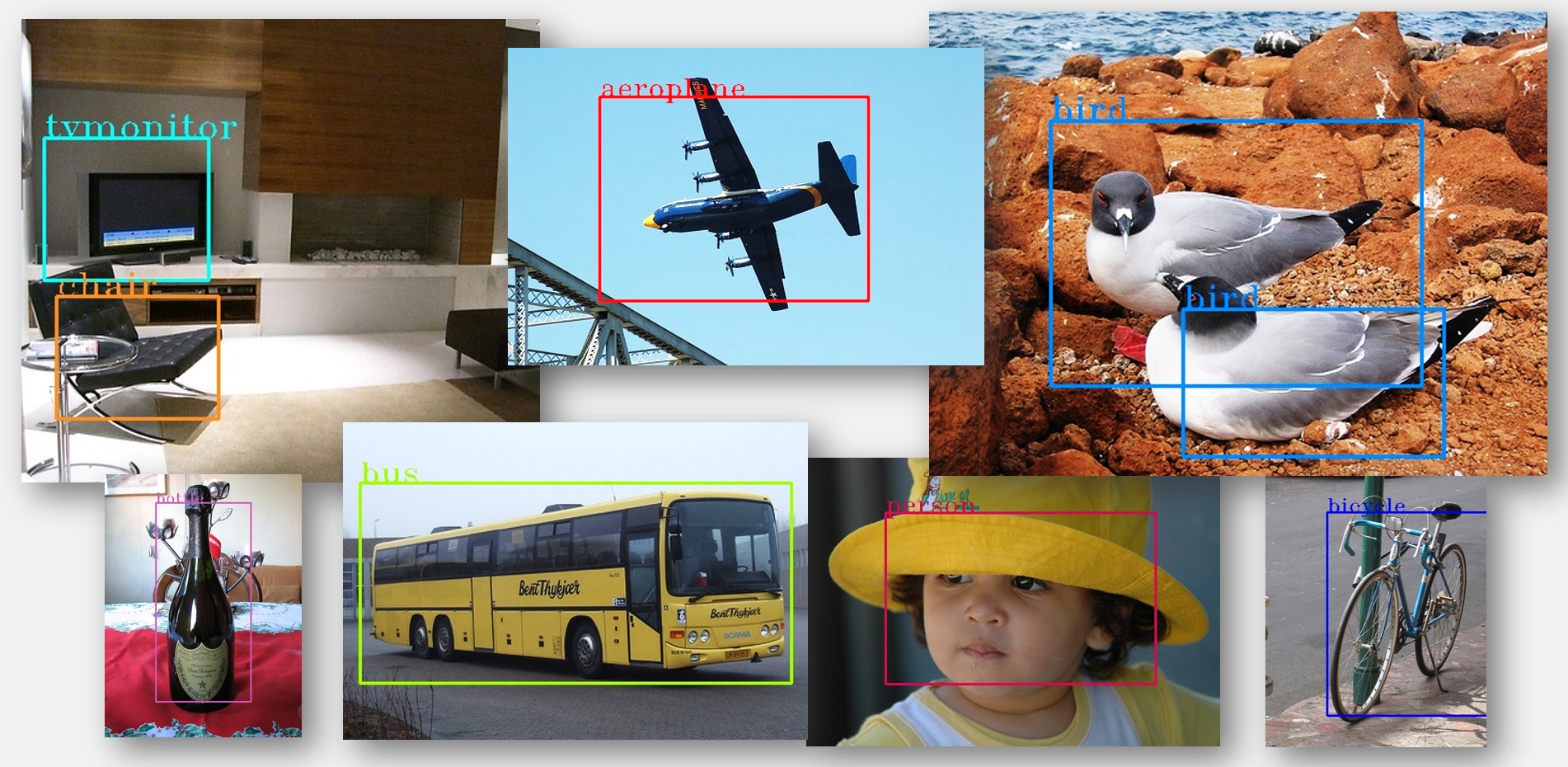}
  \caption{Visual Inspection Part 2}
\end{figure}
Overall, I think the results are considered good, given the time and resource limitation for training.

\section{Limitations and Future Improvements}
Obviously the existing limitation of original YOLO still remains and I claim no exceptional improvement over it.
This entire work is intended for self education of YOLO object detection fundamentals and familularity implementation in PyTorch.
Further improvements could be done with regards to training data to incorpoarate better dataset than VOC.
Other things to consider is experimenting with layer fusion as well as model pruning to make it super light on low powered devices.
Performance testing on various GPUs and edge devices should be done to provide validity of the network.
Also, I'm not certain about the generalization of the modified model. 

\section{Conclusion}
This was a comprehensive work that resulted in a deeper understanding of the fundamentals of YOLO and object detection in general.
Things that I have learned are the following:\\
Understanding YOLO paper with existing implementation to be able to break it down into smaller parts and digest it it step by step.
Implementing Object Detection Training/Inference Pipeline in PyTorch which includes modules such as custom dataset, custom transforms, custom loss, training and validation loops, visualization using Tensorboard
and detector module for inference\\

Overall, learning the fundamentals of real time object detection is valuable knowledge for my future work with regards to real-time instance segmentation, which was my primary motive behind this work.

\vspace{12pt}
\color{red}
\end{document}